\documentclass[11pt]{article}

\usepackage[final]{acl}

\usepackage{times}
\usepackage{latexsym}
\usepackage{float}

\usepackage[T1]{fontenc}

\usepackage[utf8]{inputenc}

\usepackage{microtype}

\usepackage{inconsolata}
\usepackage{graphicx}
\usepackage{amsmath}
\usepackage{amssymb}
\usepackage{tabularx}
\usepackage{makecell}
\usepackage{threeparttable}
\usepackage{multirow}
\usepackage{booktabs}
\usepackage{soul, color, xcolor}
\usepackage{ragged2e}
\usepackage{url}
\usepackage{colortbl}
\usepackage{cuted}

\title{\textsc{One2set} + Large Language Model: Best Partners for Keyphrase Generation}

\author{
    Liangying Shao\textsuperscript{1,2\footnotemark[1]}~
    Liang Zhang\textsuperscript{1\footnotemark[1]}~
    Minlong Peng\textsuperscript{3}~
    Guoqi Ma\textsuperscript{1}~\\
    \textbf{Hao Yue\textsuperscript{1}~
    Mingming Sun\textsuperscript{4}~
    Jinsong Su\textsuperscript{1,2\footnotemark[2]}}\\
    \textsuperscript{1}School of Informatics, Xiamen University, China\\
    \textsuperscript{2}Key Laboratory of Digital Protection and Intelligent Processing \\of Intangible Cultural Heritage of Fujian and Taiwan (Xiamen University), \\Ministry of Culture and Tourism, China \\
    \textsuperscript{3}Cognitive Computing Lab, Baidu Research, China \\
    \textsuperscript{4}Beijing Institute of Mathematical Sciences and Applications, China \\
    \texttt{\{liangyingshao,lzhang\}@stu.xmu.edu.cn},~~~\texttt{jssu@xmu.edu.cn}  \\
    \thanks{*Equal contribution.}
    \thanks{\dag Corresponding author.}
    \thanks{This work is done when Liangying Shao was interning at Cognitive Computing Lab, Baidu Research, China.}
}
\makeatletter
\def\thanks#1{\protected@xdef\@thanks{\@thanks
  \protect\footnotetext{#1}}}
\makeatother

\begin{document}
\maketitle
\begin{strip}
\vspace{-1.3cm}\quad
\end{strip}
\begin{abstract}
Keyphrase generation (KPG) aims to automatically generate a collection of phrases representing the core concepts of a given document.
The dominant paradigms in KPG include \textsc{one2seq} and \textsc{one2set}. Recently, there has been increasing interest in applying large language models (LLMs) to KPG. 
Our preliminary experiments reveal that it is challenging for a single model to excel in both recall and precision. Further analysis shows that: 1) the \textsc{one2set} paradigm owns the advantage of high recall, but suffers from improper assignments of supervision signals during training; 2) LLMs are powerful in keyphrase selection, but existing selection methods often make redundant selections. 
Given these observations, we introduce a \textit{generate-then-select} framework decomposing KPG into two steps, where we adopt a \textsc{one2set}-based model as \emph{generator} to produce candidates and then use an LLM as \emph{selector} to select keyphrases from these candidates. 
Particularly, we make two important improvements on our generator and selector: 1) we design an Optimal Transport-based assignment strategy to address the above improper assignments; 2) we model the keyphrase selection as a sequence labeling task to alleviate redundant selections.
Experimental results on multiple benchmark datasets show that our framework significantly surpasses state-of-the-art models, especially in absent keyphrase prediction. We release our code at  \url{https://github.com/DeepLearnXMU/KPG-SetLLM}.

\end{abstract}

\section{Introduction}


The keyphrase generation (KPG) task involves creating a set of phrases to encapsulate the core concepts of given document.
High-quality keyphrases enhance various downstream tasks, such as information retrieval \cite{DBLP:conf/ijcnlp/KimKCOPS13, DBLP:conf/cikm/TangHLTWYZ17, DBLP:conf/acl/BoudinGA20}, text summarization \cite{DBLP:conf/acl/WangC13, DBLP:conf/naacl/PasunuruB18}.
In general, keyphrases are categorized into two types: 1) \textit{present keyphrases} that occur continuously in the given document, and 2) \textit{absent keyphrases} that do not match any continuous subsequence. The quality evaluation of keyphrases includes two aspects: \textit{precision}, which requires the generated keyphrases to be pertinent to the document, and \textit{recall}, which demands the generated keyphrases cover the core ideas of the document.

Dominant paradigms for KPG include \textsc{one2seq} \cite{DBLP:conf/acl/YuanWMTBHT20} and \textsc{one2set} \cite{DBLP:conf/acl/YeGL0Z20}. 
The former treats KPG as a sequence generation task, while the latter treats it as a set generation by introducing multiple control codes for parallel keyphrase generation and dynamically assigning keyphrase ground-truths to control codes as supervision based on bipartite matching \cite{DBLP:books/daglib/p/Kuhn10}.
Recently, pre-trained language models (PLMs) have been widely incorporated into KPG via \textsc{one2seq} paradigm \cite{DBLP:journals/corr/abs-2201-05302, DBLP:conf/emnlp/ZhaoYYY22, DBLP:conf/emnlp/WuAC23, DBLP:conf/mm/DongWMZWLS23}.
Particularly, with the emergence of LLMs, researchers have also begun to introduce LLMs into KPG via in-context learning \cite{DBLP:journals/corr/abs-2303-13001, DBLP:journals/corr/abs-2304-14177}. 
However, it is difficult for a single model to achieve high performance in precision and recall simultaneously. 
As verified by our preliminary study (See Section \ref{study:P-R}),
models that excel in recall tend to have lower precision, while models with high precision fall short in recall. 

In this paper, to deal with the above issue, we introduce a \textit{generate-then-select} framework that decomposes the KPG task into two steps, each handled by a separate sub-model. This framework includes a generator that aims to recall correct keyphrases and a selector that eliminates incorrect candidates.

To identify the most suitable models for the generator and selector, we conduct further experiments in the preliminary study to investigate the potential of conventional KPG models and LLMs for these roles.
Our experimental results lead to two conclusions: 1) \textsc{SetTrans}, a \textsc{One2Set}-based KPG model,  has a significant advantage in recall and thus is well-suited as the generator, and 2) LLMs with their superior semantic understanding, are more effective than small language models (SLMs) for keyphrase selection and are suitable as the selector.

Furthermore, we improve the generator and selector of our framework in two aspects.
The generator assigns each ground-truth to only one control code. However, the number of control codes generally exceeds that of ground-truths, leading to insufficient training for many control codes.
To address the above improper assignments, we propose an OT-based assignment strategy for \textsc{one2set}. This strategy converts the matching of candidates and ground truth into an OT problem, allowing a ground-truth to be assigned to multiple candidates.
As for the selector, existing studies \cite{DBLP:conf/acl/KongZCLQSB23,DBLP:conf/emnlp/ChoiGKKC23,DBLP:conf/emnlp/0001YMWRCYR23,wang2024rescue} employ reranking methods to individually score candidates and then select those with high scores, which, however, results in many semantically similar candidates being selected. To address this issue, we convert keyphrase reranking into an LLM-based sequence labeling task. Leveraging the long sequence modeling capability of LLMs, we feed all candidates into the selector and have it autoregressively generate decision labels indicating whether to keep or discard the corresponding candidate. In this way, we can not only reduce the decoding search space of LLMs, but also alleviate semantic repetition by enabling the selector to fully consider the correlation between the current candidate and previous selections.
Particularly, to ensure robustness to the order of candidates, we feed them into the selector in random order during instruction tuning. This encourages the selector to develop a deeper understanding of the candidates' semantics. During inference, candidates are sorted by quality for the selector to prioritize candidates more likely to be correct.


Overall, the major contributions of our work can be summarized as follows:
\begin{itemize}
\setlength{\itemsep}{3pt}
\setlength{\parsep}{3pt}
\setlength{\parskip}{3pt}
\item Our in-depth analysis reveals that achieving high precision and recall simultaneously for a single model is challenging. Moreover, we find that \textsc{SetTrans} is advantageous as a generator, while LLM excels as a selector.
\item We design an OT-based assignment strategy to refine the training of \textsc{one2set} and enhance our selector by converting keyphrase reranking into a sequence labeling task. 
\item Experimental results and in-depth analysis of several commonly-used datasets demonstrate the effectiveness of our framework, especially in absent keyphrase prediction.
\end{itemize}


\begin{figure}[t]
\centering
\footnotesize
\includegraphics[width=0.40\textwidth, trim=0 10 0 10, clip]{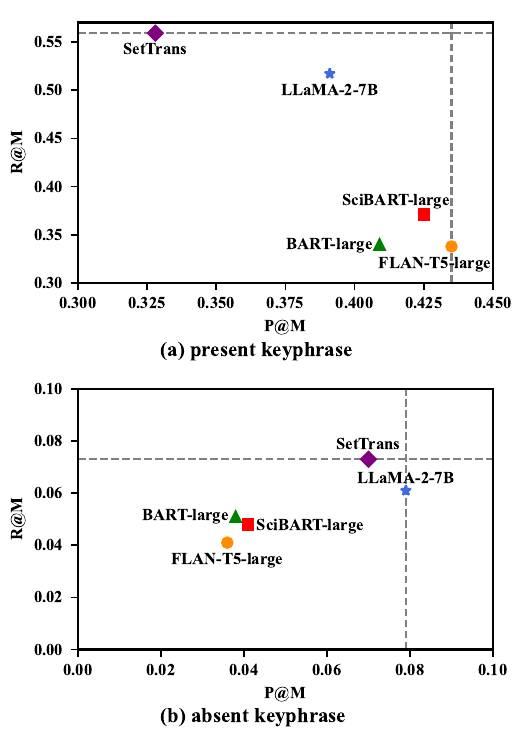}
\setlength{\abovecaptionskip}{5pt}
\caption{Performance of various models on the KP20k test set. LLaMAGen, a fine-tuned version of LLaMA-2-7B, is optimized for KPG using instruction tuning. }
\label{fig:P-R}
\vspace{-0.4cm}
\end{figure}

\section{Preliminary Study}
\label{study:P-R}

\begin{figure}[t]
\centering
\footnotesize
\includegraphics[width=0.40\textwidth, trim=5 0 5 0, clip]{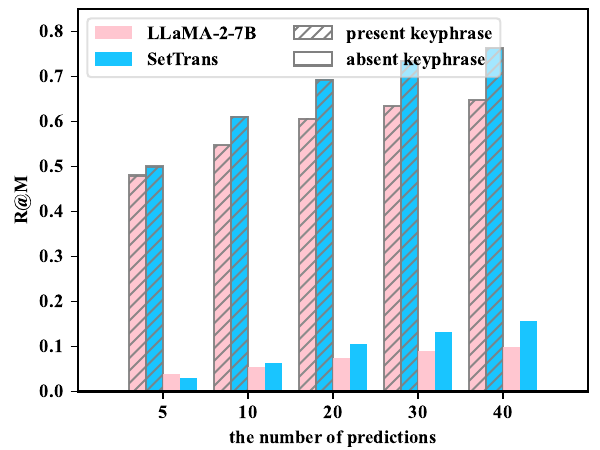}
\setlength{\abovecaptionskip}{5pt}
\caption{R@M of LLaMA-2-7B and \textsc{SetTrans} when generating the same number of keyphrases.}
\label{fig:num_recall}
\vspace{-0.4cm}
\end{figure}

To verify the necessity of decomposing KPG, we first explore the performance of dominant models in terms of recall and precision. Then, through more experiments,
we analyze which models are best suited as the generator and selector.

{\textbf{Trade-off in Keyphrase Generation.}} We measure the performance of dominant models on the testset of KP20k, including \textsc{SetTrans}, fine-tuned BART-large, SciBART-large, Flan-T5-large, and LLaMA-2-7B and report the results in Figure 1. As shown in Figure \ref{fig:P-R}, achieving high accuracy and recall simultaneously is challenging for a single model. Whether it is a conventional KPG model or an LLM, as the number of predicted keyphrases increases, the recall of the model inevitably increases while its accuracy decreases, and vice versa. This result proves the necessity of decomposing KPG into two steps.

{\textbf{Evaluating the Recall Performance of LLaMA-2-7B and \textsc{SetTrans}.}} As revealed above, \textsc{SetTrans} and LLaMA-2-7B exhibit excellent recall in commonly-used setting. In Figure \ref{fig:num_recall}, we further investigate their recall under various settings for a full comparison. We can clearly observe that as the number of generated candidates increases, \textsc{SetTrans} consistently exhibits better recall performance than LLaMA-2-7B. Given its stronger recall performance and lower computational consumption, we choose \textsc{SetTrans} as the generator.
Additionally, \textsc{SetTrans} tends to recall more correct keyphrases along with more incorrect candidates (see Appendix \ref{appendix:P-R}),  highlighting the necessity of a selector with strong filtering capabilities to improve accuracy.

{\textbf{Evaluating SLM and LLM for Keyphrase Selection.}} To identify a suitable selector, we first use \textsc{SetTrans} as the generator to output candidates, and then compare multiple representative keyphrase reranking methods. The methods we consider include 
1) SLM-Scorer, the reranker from \cite{DBLP:conf/emnlp/ChoiGKKC23}, which is an SLM-based one and achieves SOTA performance in keyphrase reranking,
and
2) LLM-Scorer, a LLaMA-2-7B reranker, which is fined-tuned as detailed in Appendix \ref{appendix:LLMScore}.
As shown in Table \ref{table:selector}, LLM-Scorer achieves higher accuracy and better F1@M scores, indicating that the powerful semantic understanding capability of LLMs is helpful for keyphrase selection. Consequently, we adopt LLaMA-2-7B as the selector.

\begin{table}[t]
\footnotesize
\centering
\renewcommand\arraystretch{1.0}
\setlength{\tabcolsep}{1.5mm}{
\begin{threeparttable}
\begin{tabular}{l|l|cc|cc}
\toprule
\multicolumn{2}{c|}
{\multirow{2}{*}{\textbf{Model}}}&
	\multicolumn{2}{c|}{{\bf Acc}}&\multicolumn{2}{c}{{\bf F1@M}}\cr
	\multicolumn{2}{c|}{}& {Pre} & {Abs} &  Pre & Abs
\cr
\midrule
\multicolumn{2}{l|}{ \textsc{SetTrans}  +  SLM-Scorer } & 0.813 & 0.806 & 0.429 & 0.082
\cr
\multicolumn{2}{l|}{ \textsc{SetTrans}  +   LLM-Scorer } & \bf 0.823 & \bf 0.812 & \bf 0.441 & \bf 0.089
\cr
\bottomrule
\end{tabular}\caption{ Keyphrase selection performance of SLM and LLM on the KP20k test set. \textbf{Acc} represents the accuracy of selection, while \textbf{Pre} and \textbf{Abs} stand for present and absent keyphrases, respectively. }
\vspace{-0.4cm}
\label{table:selector}
\end{threeparttable}}
\end{table}

\begin{figure*}[t]
\centering
\footnotesize
\includegraphics[width=0.95\textwidth, trim=50 170 50 140, clip]{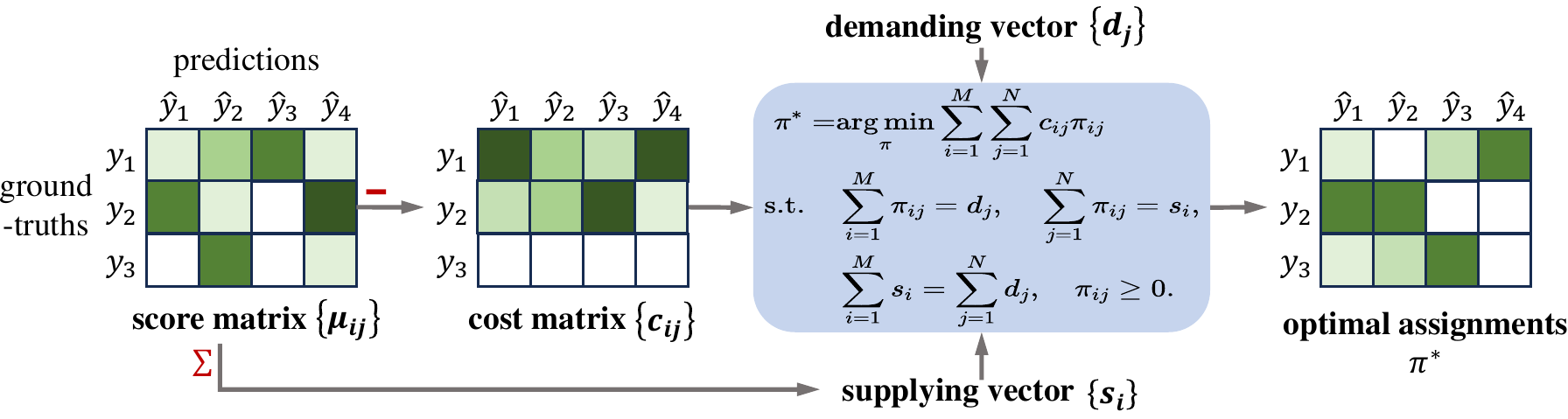}
\setlength{\abovecaptionskip}{5pt}
\caption{The OT-based supervision signal assignment for keyphrase generation. \textcolor{red}{$\sum$} represents summing as Equation \ref{formulation:supply}, while \textcolor{red}{\bf{$-$}} stands for taking the negative following Equation \ref{formulation:cost_score}. }
\label{fig:OTA}
\vspace{-0.4cm}
\end{figure*}

\section{Our Framework}
As described above, our framework involves an improved \textsc{one2set}-based generator and an LLM-based selector.
Unlike the conventional \textsc{one2set} paradigm, our generator improves the supervision signal assignment during training by modeling it as an Optimal Transport (OT) problem.
Distinct from previous studies on keyphrase reranking \cite{DBLP:conf/acl/KongZCLQSB23, DBLP:conf/emnlp/ChoiGKKC23} and LLM-based reranking \cite{DBLP:journals/corr/abs-2306-17563, DBLP:journals/corr/abs-2310-09497,ma2023coarse,wang2024rescue}, our selector autoregressively generates decision labels for keeping or discarding each candidate. This approach not only reduces the decoding search space but also fully considers the correlation between selections, thus effectively minimizing semantically repetitive selections. Moreover, we design an R-tuning S-infer strategy to help the selector comprehend the semantics of candidates.

\subsection{The \textsc{One2set}-based Generator}
\label{section:OTA}
As an extension of \textsc{SetTrans}, our generator also uses Transformer \cite{DBLP:conf/nips/VaswaniSPUJGKP17} as the backbone, of which the decoder is equipped with $N$ control codes to individually generate candidate keyphrases.
During the model training, ground-truth keyphrases $\{y_i\}_{i=1}^{M-1}$ or $\varnothing$ ($y_M$) are dynamically assigned to the control codes as supervision signals. 
Concretely, the model first predicts $K$ tokens as the prediction $\hat{y}_j$ for the $j$-th control code and then calculates a matching score $\mu_{ij}$ between the ground-truth $y_i$ and the prediction $\hat{y}_j$ via a pair-wise matching function $\mathcal{C}_{match}(*)$\footnote{The detail of $\mathcal{C}_{match}(*)$ is described in Appendix \ref{appendix:match}.}:
\begin{equation}
\setlength{\abovedisplayskip}{3pt}
    \mu_{ij} = \frac{\mathcal{C}_{match}(y_i, \hat{y}_j)^{\frac{1}{\tau}}}{\sum\limits_{j = 1}^{N}\mathcal{C}_{match}(y_i, \hat{y}_j)^{\frac{1}{\tau}}}, 
\setlength{\belowdisplayskip}{3pt}
\label{formulation:miu}
\end{equation}
where $\tau$ is a normalized hyper-parameter.

Then, instead of using bipartite matching, we consider the assignments between ground-truths and control codes as an OT problem\footnote{A detailed description of the OT problem can be found in Appendix \ref{appendix:OT}.} and search the optimal assignments with Sinkhorn-Knopp Iteration \cite{DBLP:conf/nips/Cuturi13}. Concretely, we consider the following crucial definitions in OT algorithm: 1) control codes are regarded as \textit{demanders} with a demanding vector $\{d_j\}_{j=1}^N$, where $d_j$ represents the number of ground-truths assigned to the $j$-th control code; 2) ground-truths are regarded as \textit{suppliers} with a supplying vector ${\{s_i\}}_{i=1}^M$, where $s_i$ represents the number of control codes that $y_i$ can be assigned to; 3) the cost matrix $\{c_{ij}\}_{i=1, j=1}^{M, N}$, where $c_{ij}$ represents the cost of assigning $y_i$ to the $j$-th control code. More specifically, we heuristically define them as follows:
\begin{itemize}
\setlength{\itemsep}{0pt}
\setlength{\parsep}{0pt}
\setlength{\parskip}{0pt}
\item  Since assigning multiple ground-truths to one control code at the same time may interfere with each other, we directly limit $d_j$ to 1.
\item Intuitively, if $y_i$ is highly matched with more control codes, it should be assigned to more control codes. To this end, we define $s_i$ as a dynamic number positively correlated with $\{\mu_{ij}\}_{j=1}^N$:
\begin{equation}
{ s_{i} = \begin{cases}
\lceil \sum \rm{topK}(\{\mu_{ij}\}_{j=1}^N, k)\rceil ,&\mbox{if $y_i \neq \varnothing$.}\\
N-\sum\limits_{y_{i^{\prime}} \neq \varnothing} s_{i^{\prime}}, &\mbox{otherwise.}
\end{cases}}
\label{formulation:supply}
\setlength{\belowdisplayskip}{3pt}
\end{equation}
where $\lceil \cdot \rceil$ indicates rounding up to an integer and $k$ is a predefined hyper-parameter.
\item To model the intuition that the higher the matching score between $y_i$ and $\hat{y}_j$, the lower the cost for assigning $y_i$ to the $j$-th control code, we define $c_{ij}$ as
\begin{equation}
{ c_{ij} = \begin{cases}
-\mu_{ij} ,&\mbox{if $y_i \neq \varnothing$.}\\
0 ,&\mbox{otherwise.}
\end{cases}}
\label{formulation:cost_score}
\setlength{\belowdisplayskip}{3pt}
\end{equation}
\end{itemize}

Having obtained the above vectors and matrix, we seek the optimal assignments $\pi^*$ according to the following objective function:
\begin{equation}
\setlength{\abovedisplayskip}{3pt}
\begin{aligned}
    \pi^{*} = &\mathop{\arg \min}_{\pi } \sum_{i=1}^{M} \sum_{j=1}^{N} c_{ij}\pi_{ij},\quad \pi \in \mathbb{R}^{M \times N}\\
    \text{s.t.} \quad & \sum_{i = 1}^{M} \pi_{ij} = d_j, \quad \sum_{j = 1}^{N} \pi_{ij} = s_i, \\
    &\sum_{i = 1}^{M} s_i = \sum_{j = 1}^{N} d_j, \quad \pi_{ij} \geq 0.
\end{aligned}
\setlength{\belowdisplayskip}{3pt}
\end{equation}

Finally, each control code is assigned with the ground-truth or $\varnothing$ that has the maximal assignment value as shown in the right of Figure \ref{fig:OTA}.
Note that we seek the optimal assignment plans $\pi^*_p$ for present keyphrases and $\pi^*_a$ for absent keyphrases, respectively, and subsequently calculate their cross-entropy losses accordingly.

\subsection{The LLM-based Selector}
After using the above generator to obtain candidate keyphrases, it is natural to focus on how to select high-quality keyphrases from them.
However, through in-depth analysis, we find that both traditional keyphrase reranking methods and LLM reranking methods tend to output keyphrases with serious semantic repetition.
To solve this problem, we propose to utilize LLMs to model keyphrase selection as a sequence labeling task. Furthermore, we design a random-tuning sorted-inference strategy that enables the selector to improve performance while retaining robustness to input order.

\begin{figure}[t]
\centering
\footnotesize
\includegraphics[width=0.5\textwidth, trim=120 50 110 50, clip]{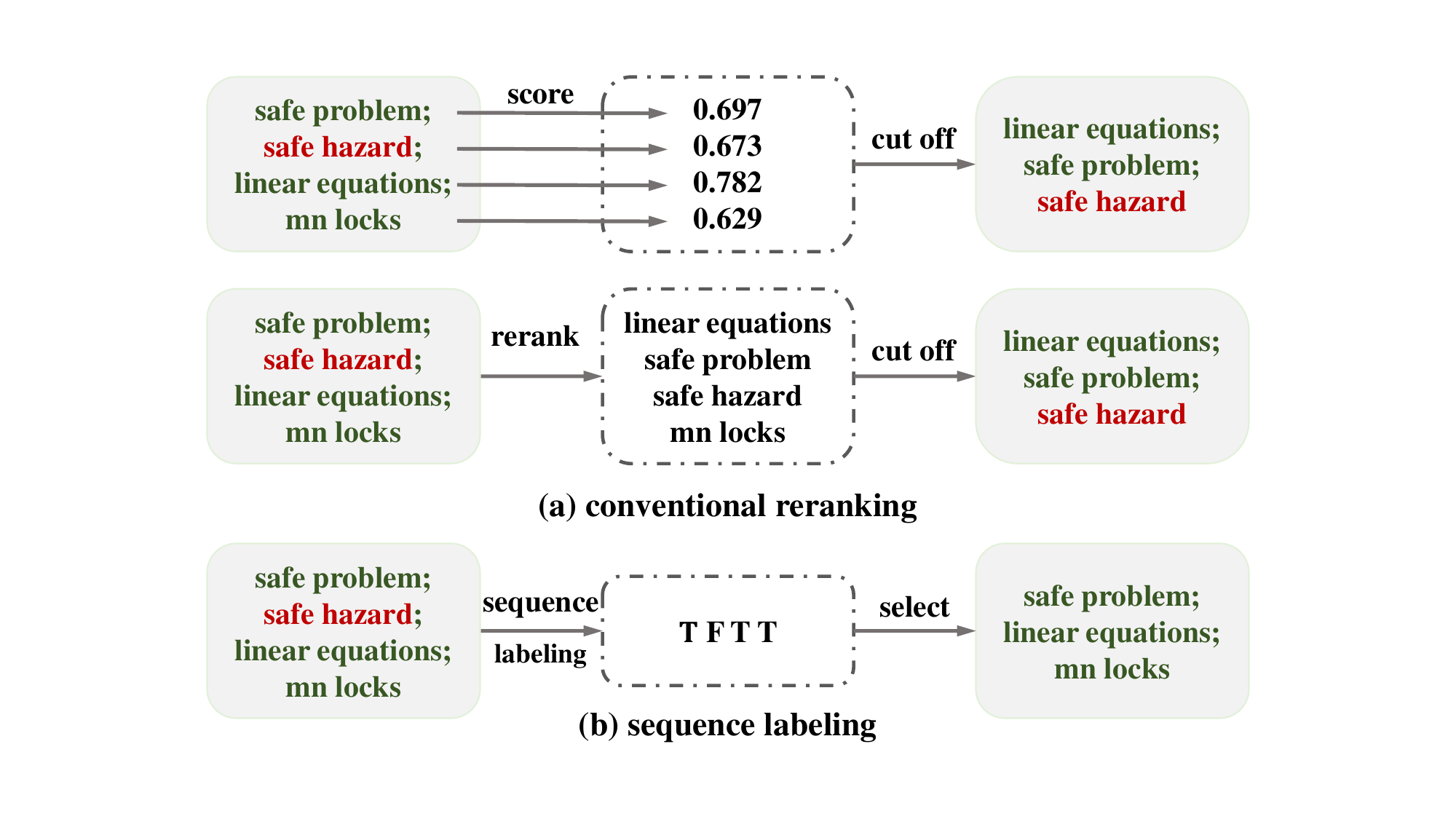}
\setlength{\abovecaptionskip}{3pt}
\caption{The comparison between the reranking methods and our sequence labeling method. A green candidate indicates a correct keyphrase, while a red candidate is an incorrect keyphrase.}
\label{fig:selector}
\vspace{-0.4cm}
\end{figure}

\paragraph{Semantic Repetition}
As shown in Figure \ref{fig:selector}(a), existing reranking studies contain the following two types: 1) one first individually score each candidate and then keep the candidates with high scores \cite{DBLP:conf/emnlp/ChoiGKKC23, DBLP:journals/corr/abs-2310-09497}, 2) the other directly ask LLMs to generate a sorted list of candidates without specific scores and save highest ranked candidates \cite{DBLP:conf/emnlp/0001YMWRCYR23, DBLP:journals/corr/abs-2306-17563}. However, when applying these methods to select keyphrases, they tend to assign similar ranks to candidates with similar semantics but different surface representations (i.e. \textit{``safe problem''} and \textit{``safe hazard''}). 

\paragraph{LLM-based Sequence Labeling}
To address the above-mentioned issue, we fine-tune an LLM to model keyphrase reranking as a sequence labeling task.
Concretely, we input all candidates into the selector and then ask it to autoregressively output decision labels, each indicating whether to keep or discard the corresponding candidate. As shown in Figure \ref{fig:selector}(b), each kept candidate is mapped to the label \textit{``T''} while each discarded one to \textit{``F''}. The instruction template we use is shown as follows:

\setlength{\parskip}{5pt}
\begin{center}
    \parbox{7cm}{
    \justifying \noindent \#\#\# \textbf{Task Definition:} 
    
    \noindent You are required to perform a sequence labeling task to select multiple keyphrases from the numbered candidates according to the given document. Use the label \textit{``T''} to indicate the selection of a candidate and the label \textit{``F''} to indicate its rejection. For instance, a label sequence \textit{``T F F''} denotes selecting candidate [1] and rejecting candidates [2] and [3].
    
    \noindent \#\#\# \textbf{Input:} 
    
    \noindent Document: $\{document\}$

    \noindent Candidates: 
    
    \noindent [1] $\{candidate_1\}$
    
    \quad ......
    
    \noindent [n] $\{candidate_n\}$
    
    \noindent \#\#\# \textbf{Response:}
    
    \noindent Label sequence: $\{label\_sequence\}$
}
\end{center}

During autoregressive generation, the selector considers previous selections when deciding whether to keep or discard the current candidate. This approach not only reduces the decoding search space but also alleviates semantic repetition.
In the example in Figure \ref{fig:selector}(b), the selector is able to discard \textit{``safe hazard''} after keeping \textit{``safe problem''}.


\paragraph{R-tuning S-infer Strategy}
Intuitively,
sorting candidates in a fixed order is beneficial for humans to select candidates.
However, such sorting may cause the selector to select candidates based on their input order rather than truly understanding their semantics.
To address this issue,
we propose a \emph{ R-tuning S-infer} strategy to handle the candidates differently during the selector training and inference.
Specifically, 
during instruction tuning,
candidates are input into the selector in a random order, encouraging the selector to learn the semantics of candidates instead of order.
By contrast,
during inference,
candidates are sorted by their quality measured with the average log probability of the generator.
In this way, the selector can prioritize candidates more likely to be correct.

\subsection{Two-stage Training}

We adopt a two-stage training strategy to train our framework, where the generator is first trained with the keyphrase generation data, and the selector is then trained with the instruction data. 

\paragraph{Generator Training} 
With the optimal assignment plans $\pi^*_p$ and $\pi^*_a$ (See Section \ref{section:OTA}), we compute the cross-entropy losses for absent and present keyphrases, respectively, and combine these two losses with weighted summation to get the final loss. Please refer to Appendix \ref{appendix:set_loss} for details.

\paragraph{Selector Training} 
When training the selector, we adopt the next-token-prediction task that has been widely used in LLMs.
Particularly, due to the imbalanced numbers of positive and negative candidates, we design the following loss:
\begin{equation}
\begin{aligned}
\setlength{\abovedisplayskip}{3pt}
\footnotesize
\mathcal{L}{(\phi)} = & \frac{1}{N_{T}}\sum_{t=1}^{|Y|} \mathbb{I}_{\{Y_t=T\}} \log p_\phi(Y_t | X, Y_{<t}) + \\
& \frac{1}{N_{F}}\sum_{t=1}^{|Y|} \mathbb{I}_{\{Y_t=F\}} \log p_\phi(Y_t | X, Y_{<t}),
\label{formulation:selector_loss}
\setlength{\belowdisplayskip}{0pt} 
\end{aligned}
\end{equation}
where $\phi$ represents the parameters of the selector, $N_{T}$ and $N_{F}$ are the numbers of \textit{``T''} and \textit{``F''}, respectively, $Y$ is the label sequence consisting of $N_{T}$ \textit{``T''} and $N_{F}$ \textit{``F''}, and $X$ is the input excluding label sequence. The negative effect of label imbalance in tuning can be mitigated by averaging loss items with identical labels.

In our experiments, we adopt QLoRA \cite{DBLP:conf/nips/DettmersPHZ23} to perform quantization on model parameters for efficient training. As such, the trainable parameters in our model are about 0.06\% of the original size.

\section{Experiments}
\subsection{Setup}
\label{section:setting}

\noindent{\textbf{Datasets.}} Following previous studies \cite{DBLP:conf/acl/MengZHHBC17, DBLP:conf/acl/YeGL0Z20, DBLP:conf/emnlp/ChoiGKKC23}, we use the training set of KP20k to train all models and then evaluate them on five benchmarks\footnote{https://huggingface.co/memray}: Inspec \cite{DBLP:conf/emnlp/Hulth03}, Krapivin \cite{krapivin2009large}, NUS \cite{DBLP:conf/icadl/NguyenK07}, SemEval \cite{DBLP:conf/semeval/KimMKB10}, KP20k \cite{DBLP:conf/acl/MengZHHBC17}.

\noindent{\textbf{Baselines.}} We compare our framework with three kinds of baselines: 1) \emph{Generative Models}: these models predict both present and absent keyphrases through generation. We consider following representative models, \textbf{catSeq} \cite{DBLP:conf/acl/YuanWMTBHT20} under \textsc{one2seq} along with its variant \textbf{ExHiRD-h} \cite{DBLP:conf/acl/ChenCLK20}, and \textbf{SetTrans} \cite{DBLP:conf/acl/YeGL0Z20} under \textsc{one2set} along with its variant \textbf{WR-one2set} \cite{DBLP:conf/emnlp/XieWYLXWZS22}. Besides, since PLM has been widely applied in KPG, we also consider two competitive models, \textbf{CorrKG} \cite{DBLP:conf/emnlp/ZhaoYYY22} and \textbf{SciBART-large + TAPT + DESEL} \cite{DBLP:conf/emnlp/WuAC23}.
2) \emph{Unified Models}: these models integrate extractive and generative methods to predict keyphrases. We report the performance of the representative models including \textbf{SEG-Net} \cite{DBLP:conf/acl/Ahmad0LC20}, \textbf{UniKeyphrase} \cite{DBLP:conf/acl/WuLLNCZW21}, \textbf{PromptKP} \cite{DBLP:conf/aaai/WuMLCN22} and \textbf{SimCKP} \cite{DBLP:conf/emnlp/ChoiGKKC23}. 3) \emph{Composite Models}: We additionally select several representative models combined like our framework.

\noindent{\textbf{Evaluation Metrics.}} 
As implemented in previous studies \cite{DBLP:conf/acl/ChanCWK19, DBLP:conf/emnlp/ZhaoYYY22, DBLP:conf/emnlp/ChoiGKKC23}, 
We evaluate all models using macro-average F1@M, and further provide the F1@5 results in Appendix \ref{appendix:F5}. Both predictions and ground-truths are stemmed with the Porter Stemmer \cite{DBLP:journals/program/Porter06}, and then the duplicates are removed before scoring. 

\noindent{\textbf{Implementation Details.}} We separately use Transformer-base \cite{DBLP:conf/nips/VaswaniSPUJGKP17} and LLaMA-2-7B \cite{DBLP:journals/corr/abs-2307-09288} to construct the generator and selector, both are optimized with Adam optimizer \cite{DBLP:conf/iclr/LoshchilovH19}. 
    
When constructing the generator, we select the top 50,002 frequent tokens to build the vocabulary. To ensure the consistency with \cite{DBLP:conf/acl/YeGL0Z20, DBLP:conf/emnlp/XieWYLXWZS22}, the number of control codes $N$ is 20, $K$ is 2, learning rate is 0.0001, and batch size is 12.
Through grid search in Appendix \ref{appendix:tau}, we set the following hyper-parameters in OT-based assignment:  $\tau=10$ in Equation \ref{formulation:miu} and Top-$3$ in Equation \ref{formulation:supply}.
During inference, we employ beam search with beam size = 10 and save all candidates for the subsequent selection.\footnote{The experiments on the impact of different beam sizes during inference are presented in Appendix \ref{appendix:beam}}

As for the selector, we adopt QLoRA with $r$ = 8, $\alpha$ = 32, and dropout of 0.05. Note that, due to the significant performance gap between present keyphrases and absent keyphrases, we use the same instruction template to tune a LoRA module for each type of keyphrase. 
The LoRA is optimized with a learning rate of 3e-4 for absent keyphrase, a learning rate of 1e-4 for present keyphrase, per-gpu batch size of 24, and the maximum epoch of 5. Validations are performed every 1,000 iterations for present keyphrase and 400 iterations for absent keyphrase, respectively. Early stopping is triggered if the validation performance does not improve in 5 consecutive rounds. 
We save the model with the best F1@M score on validation set for testing.
Particularly, we perform experiments with three random seeds and report the average results.

\subsection{Main Results}
\begin{table*}[ht]
\footnotesize
\centering
\setlength{\tabcolsep}{0.8mm}{
\begin{threeparttable}[width=0.9\textwidth]
\begin{tabular}{c|c|cc|cc|cc|cc|cc}
\toprule
\multicolumn{2}{c|}
{\multirow{2}{*}{\textbf{Model}}}&
				\multicolumn{2}{c|}{{\bf Inspec}}&\multicolumn{2}{c|}{{\bf Krapivin}}&\multicolumn{2}{c|}{{\bf NUS}}&\multicolumn{2}{c|}{{\bf SemEval}}&\multicolumn{2}{c}{{\bf KP20k}}\cr
				\multicolumn{2}{c|}{}& Pre & Abs& Pre & Abs& Pre & Abs & Pre & Abs & Pre & Abs  \cr
\midrule	\multicolumn{12}{c}{\textit{Unified Models}}\cr
\midrule
\multicolumn{2}{l|}{SEG-Net \cite{DBLP:conf/acl/Ahmad0LC20}} &0.265&0.015&0.366&0.036&0.397&0.036&0.283&0.030&0.367&0.036  \cr
\multicolumn{2}{l|}{UniKeyphrase \cite{DBLP:conf/acl/WuLLNCZW21}} &0.288&0.036&\ \ \ ---&\ \ \ ---&0.443&0.056&0.322&{0.052}&0.352&0.068  \cr
\multicolumn{2}{l|}{PromptKP \cite{DBLP:conf/aaai/WuMLCN22}} &0.294&0.022&\ \ \ ---&\ \ \ ---&0.439&0.042&0.356&0.032&0.355&0.042  \cr
\multicolumn{2}{l|}{SimCKP \cite{DBLP:conf/emnlp/ChoiGKKC23}} &0.358&0.035&0.405&{0.089}&{0.498}&{0.088}&{0.386}&0.047&0.427&{0.080}  \cr

\midrule	\multicolumn{12}{c}{\textit{Generation Models}}\cr
\midrule
\multicolumn{2}{l|}{catSeq \cite{DBLP:conf/acl/YuanWMTBHT20}} &0.262&0.008&0.354&0.036&0.397&0.028&0.283&0.028&0.367&0.032  \cr
\multicolumn{2}{l|}{ExHiRD-h \cite{DBLP:conf/acl/ChenCLK20}} &0.291&0.022&0.347&0.043&\ \ \ ---&\ \ \ ---&0.335&0.025&0.374&0.032  \cr
\multicolumn{2}{l|}{SetTrans \cite{DBLP:conf/acl/YeGL0Z20}} &0.324&0.034&0.364&0.073&0.450&0.060&0.357&0.034&0.392&0.058 \cr
\multicolumn{2}{l|}{WR-\textsc{One2Set} \cite{DBLP:conf/emnlp/XieWYLXWZS22}} &0.351&0.034&0.362&0.074&0.452&0.071&0.370&0.043&0.378&0.064 \cr
\multicolumn{2}{l|}{CorrKG \cite{DBLP:conf/emnlp/ZhaoYYY22}} &{0.365}&{0.045}&\ \ \ ---&\ \ \ ---&0.449&0.079&0.359&0.044&0.404&0.071 \cr
\multicolumn{2}{l|}{\makecell[l]{SciBART-large \cite{DBLP:conf/emnlp/WuAC23}}} &0.328&0.026&0.329&0.056&0.421&0.050&0.304&0.033&0.396&0.057 \cr
\multicolumn{2}{l|}{\quad{+ TAPT + DESEL}}&\textbf{0.402}&0.036&0.352&0.086&0.449&0.068&0.341&0.040&0.431&0.076 \cr

\multicolumn{2}{l|}{\makecell[l]{LLaMA-2-7B}} &0.344	&0.038	&{0.434}	&0.087  &	{0.481}	&0.062	&0.354	&0.042  & {0.449}	&0.069  \cr
\midrule \multicolumn{12}{c}{\textit{Composite Models}}\cr
\midrule
\multicolumn{2}{l|}{\makecell[l]{SciBART-large + Our selector}} &0.352	& {0.053}	&{0.430}	&{0.098}  &	{0.521}	&0.084	&{0.392}	&0.046  & {0.445}	&0.076  \cr
\multicolumn{2}{l|}{\makecell[l]{WR-\textsc{One2Set} + Our selector}} &0.350	& {0.059}	&{0.430}	&{0.123}  &	{0.523}	&0.118	&{0.398}	&0.057  & {0.448}	&0.108  \cr
\multicolumn{2}{l|}{\makecell[l]{Our generator + SLM-scorer}} &0.350	&0.033	&0.410	&0.094  &	{0.510}	&{0.093}	&0.390	&{0.050}  & 0.429	& {0.084}  \cr
\multicolumn{2}{l|}{{Our generator + Our selector}} &
			${0.357}_2$ & $\textbf {0.064}_1$\ddag &\ddag $ \textbf {0.435}_3$\ddag & $\textbf {0.126}_3$\ddag & $\textbf {0.528}_2$\ddag & $\textbf {0.122}_3$\ddag & $\textbf {0.405}_5$\ddag & $\textbf {0.058}_4$ & $ \textbf {0.453}_1$\ddag & $\textbf {0.112}_1$\ddag \cr
			
\bottomrule
\end{tabular}
\end{threeparttable}
\caption{Testing results on all datasets. The best performance is boldfaced, while the second best is underlined. The subscript denotes the corresponding standard deviation (e.g., $0.112_1$ indicates $0.112 \pm 0.001$). 
$\ddag$ indicates significant at $p < 0.01$ over SimCKP with 1, 000 booststrap tests \cite{tibshirani1993introduction}. 
}
\label{table:main-performance}
}
\end{table*}

The comparison results on the five testsets are shown in Table \ref{table:main-performance}. As for the present keyphrase prediction, our framework significantly outperforms all baselines on all datasets, except for Inspec.
In contrast, on the absent keyphrase prediction, our framework always performs best among all models. Note that as an extension of SciBART-large, +\,TAPT\,+\,DESEL is additionally trained in OAGKX \cite{DBLP:conf/lrec/CanoB20}, which leads to its huge improvement on Inspec. Compared to other baselines, our framework still holds comparable performance on this dataset.
Our generator combined with our selector outperforms other composite models, making it the best combination.
When comparing the results of ``Our generator + SLM-scorer'' with ``Our generator + Our selector'', it becomes evident that the LLM-based selector (our selector) demonstrates powerful filtering capabilities, underscoring the semantic understanding of LLMs in keyphrase filtering.
In the comparison of results for ``* + Our selector'', the generators based on the One2Set paradigm excel at handling absent keyphrases, with our generator achieving the best performance, indicating that the OT-based assignment strategy enhances its effectiveness.

\begin{table}[t]
\footnotesize
\centering
\renewcommand\arraystretch{1.0}
\setlength{\tabcolsep}{1.5mm}{
\begin{threeparttable}
\begin{tabular}{l|l|cc|cc}
\toprule
\multicolumn{2}{c|}
{\multirow{2}{*}{\textbf{Model}}}&
	\multicolumn{2}{c|}{{\bf In-domain}}&\multicolumn{2}{c}{{\bf Out-domain}}\cr
	\multicolumn{2}{c|}{}& Pre & Abs & Pre & Abs
\cr
\midrule
\multicolumn{2}{l|}{Ours} & {\bf 0.453} & {\bf0.112} & \textbf{0.431} & \bf 0.093
\cr
\midrule
\multicolumn{2}{l|}{ \quad $\Rightarrow$bipartite matching} & 0.446 & 0.105 & 0.421 & 0.087
\cr
\multicolumn{2}{l|}{ \quad $\Rightarrow$GenKP} & 0.441 & 0.089 & 0.382 & 0.062
\cr
\multicolumn{2}{l|}{\quad $\Rightarrow$R-tuning R-inference} & {0.442} & {0.102} & {0.416} &  {0.083}
\cr
\multicolumn{2}{l|}{ \quad $\Rightarrow$S-tuning R-inference } & 0.264 & 0.060 & 0.243 & {0.048}
\cr
\multicolumn{2}{l|}{ \quad $\Rightarrow$CE\_loss } & 0.435 & 0.096 & 0.412 & {0.062}
\cr
\bottomrule
\end{tabular}\caption{Ablation study. $\Rightarrow$* means replacing the corresponding component of our framework with *. We use three random orders and report the average performance in the R-inference setting.
}
\label{table:Ablation}
\end{threeparttable}}
\end{table}
\subsection{Ablation Study}
In Table \ref{table:Ablation}, we investigate the effect of each component on our framework to verify their validity. Following previous studies \cite{DBLP:conf/emnlp/XieWYLXWZS22, DBLP:conf/emnlp/ChoiGKKC23}, we conduct experiments on two kinds of test sets: 1) \textbf{in-domain}, which is KP20k, and 2) \textbf{out-of-domain}, which is the combination of Inspec, Krapivin, NUS, and SemEval.

\begin{figure*}[t]
\centering
\footnotesize
\includegraphics[width=0.90\textwidth, trim=15 0 15 5, clip]{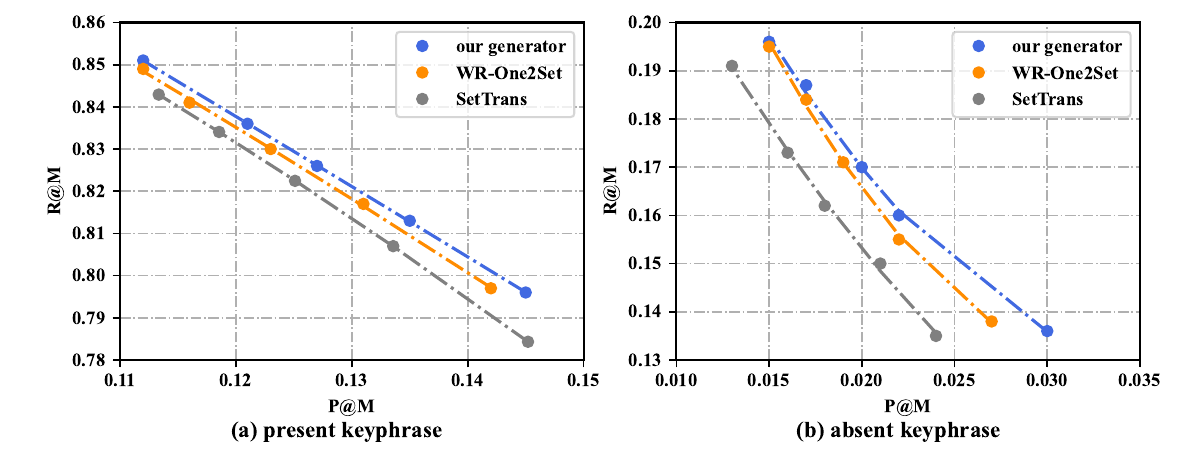}
\setlength{\abovecaptionskip}{5pt}
\caption{The recall and precision of our generator and other baselines.}
\label{fig:OTA_ablation}
\end{figure*}

(1) \textit{$\Rightarrow$bipartite matching}. In this variant, we replace the OT-based assignment with bipartite matching and observe the performance degradation in both present and absent keyphrases. Furthermore, we compare the recall scores of our generator, \textsc{SetTrans}, and WR-\textsc{One2Set} at the same precision level. From Figure \ref{fig:OTA_ablation}, the recall of our generator consistently exceeds those of \cite{DBLP:conf/acl/YeGL0Z20} and \cite{DBLP:conf/emnlp/XieWYLXWZS22} with close accuracy. 
Both experiments demonstrate the effectiveness of our OT-based assignment.

(2) \textit{$\Rightarrow$GenKP}. In this variant, we tune the selector to generate the list of kept candidates directly.
Compared to our selector, its prediction performance significantly drops, especially on out-of-domain datasets.
We argue that the sequence labeling task adopted by our selector reduces the decoding space, thus effectively reducing the task difficulty and improving generalization. 

(3) $\Rightarrow$\textit{R-tuning R-inference}. Unlike our selector, this variant inputs candidates into the selector in a random order during both training and inference, resulting in a slight performance drop. This indicates that our sorted-inference strategy helps the selector make better selections.

(4) $\Rightarrow$\textit{S-tuning R-inference}. Different from the above variant, this variant inputs candidates into the selector in a fixed order during training while in a random order during inference. Compared to R-tuning R-inference, we observe a more significant performance degradation, suggesting that the random-tuning strategy enhances the robustness of the selector to the input order.

(5) $\Rightarrow$\textit{CE\_loss}. In this variant, we tune the selector with vanilla cross-entropy loss without loss item averaging. The removal of loss item averaging notably diminishes the performance of the selector, demonstrating the effectiveness of this operation.
The fact that there are more incorrect candidates than correct ones leads to the overfitting of incorrect candidates when training with vanilla cross-entropy loss.

\subsection{Diversity of Predicted Keyphrases}
Following \citet{DBLP:journals/corr/abs-2303-15422}, we take $emb\_sim$ and $dup\_token\_ratio$ as the diversity metrics. 
As shown in Table \ref{table:diversity}, the semantic repetition in the original candidate set is severe but significantly reduced by selection models. Among these methods, our selector obtains the lowest $emb\_sim$ and $dup\_token\_ratio$, demonstrating its effectiveness in reducing semantic repetition. 

\noindent{Please see \textbf{Appendix~\ref{appendix:LLM}} for more experiments}.
\begin{table}[t]
\footnotesize
\centering
\setlength{\tabcolsep}{2.0mm}{
\begin{threeparttable}
\begin{tabular}{l|l|c|c}
\toprule
\multicolumn{2}{c|}
{\multirow{1}{*}{\textbf{Model}}} & 
\multicolumn{1}{c|}{\textbf{Dup\_token\_ratio} $\downarrow$} & 
\multicolumn{1}{c}{{\bf emb\_sim $\downarrow$}}
\cr
\midrule
\multicolumn{2}{l|}{Our generator} & 0.406 & 0.198 \cr 
\multicolumn{2}{l|}{\quad + SLM-Scorer } & 0.330 & 0.155 \cr 
\multicolumn{2}{l|}{\quad + LLM-Scorer } & 0.247 & 0.149 \cr
\multicolumn{2}{l|}{\quad + Our Selector } & \bf 0.222 & \bf 0.147 \cr
\midrule
\multicolumn{2}{l|}{ ground-truth } &  0.072 &  0.132 \cr

\bottomrule
\end{tabular}
\caption{
The diversity of all keyphrases produced by various models. 
\label{table:diversity}}
\setlength{\belowcaptionskip}{0pt}
\end{threeparttable}}
\vspace{-0.2cm}
\end{table}

\section{Related Work}
The related works to ours mainly include keyphrase generation and keyphrase selection.

{\textbf{Keyphrase Generation.}}
Generally, KPG models are constructed under the following paradigms: 
1) \textsc{one2one} \cite{DBLP:conf/acl/MengZHHBC17}, where keyphrases of each document are split and each keyphrase along with the document forms a training instance. During inference, top-$K$ candidates are picked under beam search.
2) \textsc{one2seq} \cite{DBLP:conf/acl/YuanWMTBHT20}, which treats KPG as a sequence generation task, concatenating keyphrases into a sequence according to a predefined order. 
3) \textsc{one2set} \cite{DBLP:conf/acl/YeGL0Z20}, which generates keyphrases as an unordered set conditioned on learnable control codes. 
Among these paradigms, \textsc{one2set} excels in recall. \citet{DBLP:conf/acl/YeGL0Z20} utilize the bipartite matching to assign ground-truths or $\varnothing$ to control codes as the supervision signal. 
Furthermore, \citet{DBLP:conf/emnlp/XieWYLXWZS22} propose a re-assignment mechanism to refine the assignment results of the bipartite matching, which allows a proportion of control codes matched with $\varnothing$ to learn ground-truths. 


{\textbf{Keyphrase Selection.}}
Currently, researchers \cite{DBLP:conf/emnlp/SongJX21, DBLP:conf/acl/ZhangCWDZLW022, DBLP:conf/acl/KongZCLQSB23} select keyphrases from candidates using reranking methods. The common practice is to perform phrase mining on n-grams within document to extract candidates. Recently, \citet{DBLP:conf/emnlp/ChoiGKKC23} obtain present keyphrase candidates through data mining and absent keyphrase candidates from the KPG model. All the above methods individually score each candidate with PLMs and select those with high scores. However, this independent scoring leads to semantic repetition issue.
With the rapid development of LLMs, researchers try to insert the candidates into prompt and instruct the LLMs to generate an ordered list \cite{DBLP:conf/emnlp/SachanLJAYPZ22, DBLP:conf/emnlp/0001YMWRCYR23, DBLP:journals/corr/abs-2310-09497, DBLP:journals/corr/abs-2306-17563,ma2024large}, which demonstrates impressive effectiveness in document reranking tasks. 

Overall, our work differs from previous studies for two main reasons.
First, unlike \cite{DBLP:conf/emnlp/XieWYLXWZS22}, we treat the matching of control codes and ground-truths as an OT problem and propose an OT-based assignment strategy to refine the target assignment in the \textsc{one2set} paradigm. Second, in contrast to current selection methods, we consider keyphrase selection as an LLM-based sequence labeling task, where the correlation between the current candidate and previous selections can be fully exploited.

\section{Conclusion and Future Work}
\label{sec:bibtex}

This paper introduces a generate-then-select framework that integrates a \textsc{one2set} model and an LLM selector together, so as to fully leverage the high recall of the \textsc{one2set} paradigm and powerful semantic understanding of LLM. The \textsc{one2set} model acts as the generator and is optimized by our OT-based assignment to recall more correct candidates. The LLM acts as the selector that models the selection of keyphrase candidates as a sequence labeling task and reduces the semantic repetition through its long sequence modeling capability. Experimental results show that our framework achieves significant performance improvements compared to existing state-of-the-art models.

In the future, we tend to combine the generation and selection tasks into a multi-task learning framework, which further improves the synergy between the two tasks.

\section*{Acknowledgments}
The project was supported by National Natural Science Foundation of China (No. 62276219), and the Public Technology Service Platform Project of Xiamen (No. 3502Z20231043). We also thank the reviewers for their insightful comments.

\section*{Limitations}
While this paper introduces a generate-then-select framework for KPG that effectively combines the strengths of the \textsc{one2set} paradigm and LLM, it has several limitations in terms of resource consumption. First, the LLM is inherently resource-intensive due to its large number of parameters, demanding significant computational power and memory. Second, the two-step process of generating and then selecting keyphrases is time-consuming, which can lead to relative inefficiency in practical applications. These factors combined make the proposed framework more resource-consuming and challenging to implement compared to single-model solutions.

\bibliography{custom}

\begin{table*}[]
\vspace{-0.2cm}
\footnotesize
\centering
\setlength{\tabcolsep}{0.8mm}{
\begin{threeparttable}[width=0.9\textwidth]
\begin{tabular}{c|c|cc|cc|cc|cc|cc}
\toprule
\multicolumn{2}{c|}
{\multirow{2}{*}{\textbf{Model}}}&
				\multicolumn{2}{c|}{{\bf Inspec}}&\multicolumn{2}{c|}{{\bf Krapivin}}&\multicolumn{2}{c|}{{\bf NUS}}&\multicolumn{2}{c|}{{\bf SemEval}}&\multicolumn{2}{c}{{\bf KP20k}}\cr
				\multicolumn{2}{c|}{}& Pre & Abs& Pre & Abs& Pre & Abs & Pre & Abs & Pre & Abs  \cr
\midrule
\multicolumn{2}{l|}{SimCKP} &\bf 0.358 & 0.035 & 0.405 & 0.089 & 0.498 & 0.088 & 0.386 & 0.047 & 0.427 & 0.080  \cr
\multicolumn{2}{l|}{Our generator + Our selector} &0.357 & \bf 0.064 & \bf 0.435 & \bf 0.126 & \bf 0.528 & \bf 0.122 & \bf 0.405 & \bf 0.058 & \bf 0.453 & \bf 0.112  \cr
\bottomrule
\end{tabular}
\end{threeparttable}
\caption{F1@5 results on all datasets.
}
\label{table:F1@5}
}
\end{table*}

\begin{table*}[]
\vspace{-0.2cm}
\footnotesize
\centering
\setlength{\tabcolsep}{0.8mm}{
\begin{threeparttable}[width=0.9\textwidth]
\begin{tabular}{c|c|cc|cc|cc|cc|cc}
\toprule
\multicolumn{2}{c|}
{\multirow{2}{*}{\textbf{Model}}}&
				\multicolumn{2}{c|}{{\bf Inspec}}&\multicolumn{2}{c|}{{\bf Krapivin}}&\multicolumn{2}{c|}{{\bf NUS}}&\multicolumn{2}{c|}{{\bf SemEval}}&\multicolumn{2}{c}{{\bf KP20k}}\cr
				\multicolumn{2}{c|}{}& Pre & Abs& Pre & Abs& Pre & Abs & Pre & Abs & Pre & Abs  \cr
\midrule
\multicolumn{2}{l|}{SimCKP \cite{DBLP:conf/emnlp/ChoiGKKC23}} &0.358&0.035&0.405&{0.089}&{0.498}&{0.088}&{0.386}&0.047&0.427&{0.080}  \cr
\multicolumn{2}{l|}{\makecell[l]{SciBART-large \cite{DBLP:conf/emnlp/WuAC23}}} &0.328&0.026&0.329&0.056&0.421&0.050&0.304&0.033&0.396&0.057 \cr
\multicolumn{2}{l|}{\quad{+ TAPT + DESEL}}&\textbf{0.402}&0.036&0.352&0.086&0.449&0.068&0.341&0.040&0.431&0.076 \cr
\multicolumn{2}{l|}{\makecell[l]{Zephyr-7B}} &0.358	&0.035	&{0.428}	&0.092  &	{0.479}	&0.058	&0.353	&0.047  & {0.443}	&0.072  \cr
\multicolumn{2}{l|}{Our generator + Our selector (Zephyr-7B)} &0.374 & \bf 0.058 & \bf 0.430 & \bf 0.133 & \bf 0.518 & \bf 0.114 & \bf 0.401 & \bf 0.067 & \bf 0.448 & \bf 0.114 \cr
\bottomrule
\end{tabular}
\end{threeparttable}
\caption{F1@M results on different LLM-based selectors.
}
\label{table:open-LLMs}
}
\vspace{-0.4cm}
\end{table*}

\newpage

\appendix
\begin{table*}[t]
\vspace{-0.2cm}
\footnotesize
\centering
\setlength{\tabcolsep}{1.2mm}{
\begin{threeparttable}[width=0.9\textwidth]
\begin{tabular}{c|c|cc|cc|cc|cc|cc}
\toprule
\multicolumn{2}{c|}
{\multirow{2}{*}{\textbf{Model}}}&
				\multicolumn{2}{c|}{{\bf Inspec@100}}&\multicolumn{2}{c|}{{\bf Krapivin}@100}&\multicolumn{2}{c|}{{\bf NUS@100}}&\multicolumn{2}{c|}{{\bf SemEval@100}}&\multicolumn{2}{c}{{\bf KP20k@100}}\cr
				\multicolumn{2}{c|}{}& Pre & Abs& Pre & Abs& Pre & Abs & Pre & Abs & Pre & Abs  \cr
\midrule
\multicolumn{2}{l|}{SimCKP \cite{DBLP:conf/emnlp/ChoiGKKC23}} &0.342&0.046&0.397&{0.101}&{0.482}&{0.064}&{0.387}&0.051&0.420&{0.086}  \cr
\multicolumn{2}{l|}{\makecell[l]{LLaMA-2-7B}} &0.307	&0.048	&{0.453}	&0.073  &	{0.498}	&0.042	&0.372	&0.034  & {0.413}	&0.056  \cr
\multicolumn{2}{l|}{GPT-4} & \bf 0.513 & 0.055 & 0.343 & 0.010 & 0.244 & 0.037 & 0.356 & 0.024 & 0.224 & 0.015 \cr
\multicolumn{2}{l|}{Our generator + GPT-4 selector} &0.435 & 0.067 & 0.435 & 0.077 & 0.400 & 0.077 & 0.354 & 0.034 & 0.309 & 0.038 \cr
\multicolumn{2}{l|}{Our generator + Our selector} & 0.327 & \bf 0.069 & \bf 0.457 & \bf 0.142 & \bf 0.499 & \bf 0.136 & \bf 0.405 & \bf 0.058 & \bf 0.424 & \bf 0.097 \cr
\bottomrule
\end{tabular}
\end{threeparttable}
\caption{F1@M results on GPT-4.}
\label{table:close-LLMs}
}
\vspace{-0.4cm}
\end{table*}

\section{Experimental Results of F1@5}
\label{appendix:F5}
The F1@5 results of “our generator + our selector” are shown in Table \ref{table:F1@5}. On this metric, our framework also outperforms the strongest baseline, proving its effectiveness.

\section{Further Experiments}

\subsection{Experimental Results on More LLMs}
\label{appendix:LLM}
\paragraph{More Results on Open LLMs}
As described previously, we mainly conduct experiments on LLaMA-2-7b, due to its widespread use in the research community. 
To verify the validity of our framework on other LLMs, we report the results of experiments using Zephyr-7B \cite{DBLP:journals/corr/abs-2310-16944}. 
As shown in Table \ref{table:open-LLMs}, our framework still achieves better performance to other baselines, suggesting that our framework is not sensitive to the choice of LLMs.

\paragraph{More Results on Close LLMs}
We conduct experiments to report the few-shot performance of GPT-4 as the generator and selector, respectively. The experiment results on the top-100 samples of each dataset are shown Table \ref{table:close-LLMs}.
Consistent with previous studies \cite{song2023chatgpt,martinez2023chatgpt,DBLP:journals/ipm/XieSSWWYLXS23}, 
GPT-4 also excels in the present keyphrase prediction of Inspec dataset.
However, on other datasets, GPT-4 exhibits significantly worse performance than Our generator + Our selector, highlighting the effectiveness of our framework.
Furthermore, when replacing our selector with GPT-4, Our generator + GPT-4 selector is still inferior to Our generator + Our selector on most datasets, demonstrating that One2set generator and LLM-based selector are the best combination for keyphrase generation.

\subsection{Recall and Precision of \textsc{SetTrans} with Different Numbers of Prediction}
\label{appendix:P-R}
As mentioned in Section \ref{study:P-R}, \textsc{SetTrans} tends to recall more correct keyphrases along with more incorrect candidates. We report the recall and precision of \textsc{SetTrans} in Table \ref{tab:num_R}. As the number of predicted keyphrases increases, recall greatly improves, but precision significantly drops, indicating that more incorrect candidates are generated. This underscores the urgent need for a strong selector to improve accuracy.
\begin{table}[ht]
\footnotesize
\centering
\renewcommand\arraystretch{1.0}
\setlength{\tabcolsep}{2mm}
\begin{threeparttable}
\begin{tabular}{l|cc|cc|c}
\toprule
\multirow{2}{*}{\textbf{Model}}&
	\multicolumn{2}{c|}{{\bf Present}}&
        \multicolumn{2}{c|}{{\bf Absent}}& 
        \multirow{2}{*}{\textbf{Num}} 
        \cr
 & $P@M$ & $R@M$ & $P@M$ & $R@M$ &
\cr
\midrule
\multirow{5}{*}{SetTrans} 
& 0.340     & 0.500  & 0.050     & 0.030  & 5
\cr
& 0.274     & 0.611  & 0.066     & 0.064  & 10  
\cr
& 0.221     & 0.693  & 0.044     & 0.105  & 20 
\cr
& 0.195     & 0.734  & 0.032     & 0.133  & 30 
\cr
& 0.178     & 0.763  & 0.026     & 0.156  & 40 
\cr 
\bottomrule
\end{tabular}
\caption{Recall and precision of \textsc{SetTrans} with different prediction numbers.}
\label{tab:num_R}
\end{threeparttable}
\end{table}

\subsection{Effect of $\tau$}
\label{appendix:tau}
\begin{figure}[t]
\centering
\footnotesize
\includegraphics[width=0.40\textwidth, trim=10 10 15 10, clip]{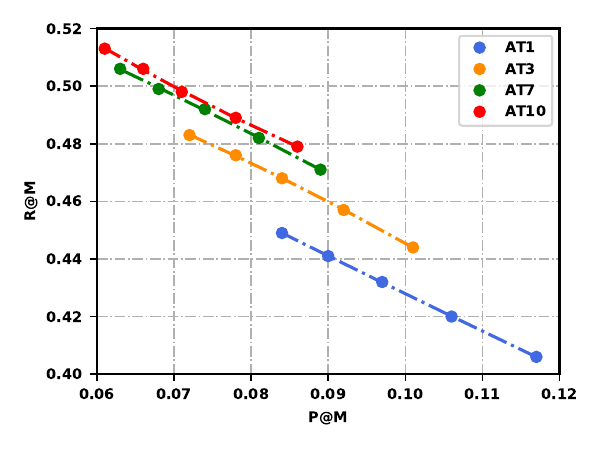}
\setlength{\abovecaptionskip}{5pt}
\caption{The recall and precision of our generator over different $\tau$ on KP20k validation set.}
\label{fig:temperature}
\vspace{-0.4cm}
\end{figure}

We conduct experiments on various $\tau$ for smoothing the matching scores among predictions and ground-truth keyphrases and report the precision and recall of the generator under various beam sizes. As shown in Figure \ref{fig:temperature}, the generator achieves the best performance at 10. Therefore, we adopt $\tau$ = 10 in all experiments.

\subsection{Effect of Different Beam Sizes}
\label{appendix:beam}


\begin{table}[t]
\footnotesize
\centering
\renewcommand\arraystretch{1.0}
\setlength{\tabcolsep}{1.2mm}
\begin{threeparttable}
\begin{tabular}{c|cc}
\toprule
\multirow{2}{*}{\textbf{Model}}&
	\multicolumn{2}{c}{{\bf $F1@M$}}
        \cr
 & Pre & Abs 
\cr
\midrule
\multicolumn{1}{l|}{Our generator ($\#bs$ = 1) + Our selector} &  0.438 & 0.050
\cr
\midrule
\multicolumn{1}{l|}{Our generator ($\#bs$ = 5) + Our selector} & \bf0.457 &  0.065
\cr
\midrule
\multicolumn{1}{l|}{Our generator ($\#bs$ = 10) + Our selector} &  \bf0.457 &  \bf 0.080
\cr
\midrule
\multicolumn{1}{l|}{Our generator ($\#bs$ = 15) + Our selector} & \bf0.457 & \bf0.080
\cr
\bottomrule
\end{tabular}
\caption{Performance on KP20k validation set. $\#bs$ denotes beam size.}
\label{tab:P-R-ablation}
\end{threeparttable}
\vspace{-0.2cm}
\end{table}

We investigate the impact of beam size on the KP20k validation set. To this end, we gradually varied beam size from 1 to 15. 
As shown in Table \ref{tab:P-R-ablation}, 
both present keyphrases and absent keyphrases achieved the best performance with a beam size of 10, and the performance is maintained as the beam size increased further. Therefore, we use a beam size of 10.

\section{Algorithm Details}

\subsection{Formulation of $\mathcal{C}_{match}$}
\label{appendix:match}
Same as \cite{DBLP:conf/acl/YeGL0Z20}, we generate $K$ tokens conditioned on each control code and collect their predictive probability distributions $\{P_j\}_{j=1, 2, ..., N}$ and $P_j = \{p^t_{j}\}_{t=1, ..., K}$, where $p_j^t$ is the predictive distribution at time step $t$ for control code $j$. The matching score between any pair of ground-truth $y_i$ and candidate $\hat{y}_{\pi(i)}$ is calculated as following:
\begin{equation}
\setlength{\abovedisplayskip}{3pt}
    \mathcal{C}_{match}(y_i, \hat{y}_{\pi(i)}) = -\sum_{t=1}^{K'}  \mathbb{I}_{\{y_i^t\neq\varnothing\}}\ p_{\pi(i)}^t(y_i^t)
\setlength{\belowdisplayskip}{3pt}
\end{equation}
where $K' = \min(\left| y_i\right |, K)$ and $p_{\pi(i)}^t(y_i^t)$ represents the probability of token $y_i^t$ in $p_{\pi(i)}^t$. The scores from matching any prediction with $\varnothing$ are set to 0, which avoids interference in the assignment of valid ground-truths.

\subsection{Optimal Transport}
\label{appendix:OT}
Assume a scenario involving $m$ suppliers and $n$ demanders, where the $i$-th supplier holds $s_i$ units of goods and the $j$-th demander needs $d_j$ units of goods. Every route between a supplier and demanders has a per-unit transportation cost, denoted by $c_{ij}$. The objective of Optimal Transport (OT) is to seek the most efficient distribution plan $\pi^* = \{\pi_{ij} | i = 1, 2, \ldots, m, j = 1, 2, \ldots, n\}$, which minimizes the total cost of transporting all goods from suppliers to demanders. 

An equivalent mathematical formulation of the OT problem is presented as follows:
\vspace{-0.2cm}
$$
\begin{aligned}
	\min_\pi \quad & \sum_{i = 1}^{m} \sum_{j = 1}^{n} c_{ij} \pi_{ij} \\
	\text{s.t.} \quad & \sum_{i = 1}^{m} \pi_{ij} = d_j, \quad \sum_{j = 1}^{n} \pi_{ij} = s_i, \\
	&\sum_{i = 1}^{m} s_i = \sum_{j = 1}^{n} d_j, \quad \pi_{ij} \geq 0, \\
	&i = 1, 2, \ldots, m,\quad  j = 1, 2, \ldots, n.
\end{aligned}
$$
\vspace{-0.2cm}

In order to tackle the OT problem efficiently, we employ the Sinkhorn-Knopp Iterative algorithm \cite{DBLP:conf/nips/Cuturi13}, which demonstrates polynomial-time complexity.

\subsection{Formulation of The Generator Loss Function}
\label{appendix:set_loss}
Following \cite{DBLP:conf/acl/YeGL0Z20}, we organize the loss of our generator as:
\begin{equation}
\setlength{\abovedisplayskip}{3pt}
\footnotesize
{\mathcal{L}(\theta) = -{\Bigg[\sum_{i=1}^{\frac{N}{2}}}{\mathcal{L}^p(\theta, y_{\pi^*_{p}(i)})+ \sum_{i=\frac{N}{2}+1}^{N}}{\mathcal{L}^a(\theta, y_{\pi^*_{a}(i)})\Bigg]}
}\label{formulation:generator_total_loss}
\setlength{\belowdisplayskip}{0pt}
\end{equation}

\begin{equation}
\footnotesize
\setlength{\abovedisplayskip}{0pt}
{\mathcal{L}^p(\theta, y_i)=\begin{cases}
\lambda_{pre}\cdot\sum_{t=1}^{|y_i|}\log\hat{p}_i^t(y_i^t), &\mbox{if $y_i$=$\varnothing$.}\\
\sum_{t=1}^{|y_i|}\log\hat{p}_i^t(y_i^t), &\mbox{otherwise}.
\end{cases}}
\label{formulation:generator_total_loss_OTA}
\setlength{\belowdisplayskip}{3pt}
\end{equation}
where $y_i^t$ is the $t$-th token of target $y_i$, $\hat{p}_i^t$ denotes the predictive probability distribution of the $i$-th candidate at $t$-th step, $\lambda_{pre}$ is a hyper-parameter used to decrease the impact of excessive $\varnothing$. $\mathcal{L}^a(\theta, y_i)$ is defined in the same form as $\mathcal{L}^p(\theta, y_i)$, except that $\lambda_{pre}$ is replaced with $\lambda_{abs}$. We adopt $\lambda_{pre}$ as 0.2 and $\lambda_{abs}$ as 0.1.

\section{Implementation Details}

\subsection{LLaMA-2-7B}
\label{appendix:LLaMAGen}
We use LLaMA-2-7B to perform instruction tuning and use the prompt template as follows:
\setlength{\parskip}{5pt}
\begin{center}
    \parbox{7cm}{
    \justifying \noindent \#\#\# \textbf{Instruction:} Generate keyphrases for the given document and use ; to space keyphrases. For example, ``$\rm phrase_A; phrase_B; phrase_C$''.
    
    \noindent \#\#\# \textbf{Input:} Document: $\{document\}$
    
    \noindent \#\#\# \textbf{Response:} Keyphrases: $\{keyphrases\}$
}
\end{center}
We adopt QLoRA and a learning rate of 2e-4. Validation is performed every 1,000 iterations. The rest of the experimental setup is consistent with \ref{section:setting}.



\subsection{LLM-Scorer}
\label{appendix:LLMScore}
We use LLaMA-2-7B to perform instruction tuning and use the prompt template as follows:
\setlength{\parskip}{5pt}
\begin{center}
    \parbox{7cm}{
    \justifying \noindent \#\#\# \textbf{Instruction:} 
    
    \noindent Score each candidate according to the given document.
    
    \noindent \#\#\# \textbf{Input:} 
    
    \noindent Document: $\{document\}$
    
    \noindent Candidates: 
    
    \noindent [1] $\{candidate_1\}$
    
    \quad ……
    
    \noindent [n] $\{candidate_n\}$

    \noindent \#\#\# \textbf{Response:} 
    
    \noindent Score: $\{scores\}$
}
\end{center}
The LLM-Scorer predicts the scores for all candidates in one step by generating a logit distribution, where $candidate_1$ maps to token ``<0x00>'', $candidate_2$ maps to token ``<0x01>'', and so forth. We extract the logits corresponding to these indices and assign these values as the scores for $candidate_1$ to $candidate_n$. The scorer is then tuned using the contrastive loss function proposed by \cite{DBLP:conf/emnlp/ChoiGKKC23}, which helps to maximize the distinction between the correct and incorrect candidates by adjusting the scores accordingly.

We adopt QLoRA and a learning rate of 3e-4 to tune a LoRA module for both present and absent keyphrases. The rest of the experimental setup is consistent with \ref{section:setting}.


\end{document}